\newif\ifanonsubmission
\documentclass[conference]{IEEEtran}
\usepackage{times}

\usepackage[numbers]{natbib}
\usepackage{booktabs}
\usepackage{multirow}
\usepackage{multicol}
\usepackage[bookmarks=true]{hyperref}
\usepackage{graphicx}
\usepackage{algorithm}
\usepackage{algpseudocode}
\usepackage{amsmath}
\usepackage{amssymb}
\usepackage{ifthen}
\usepackage{caption}

\let\oldtwocolumn\twocolumn
\renewcommand\twocolumn[1][]{%
    \oldtwocolumn[{#1}{
    \begin{center}
       \vspace{-1.cm}
       \includegraphics[width=\textwidth]{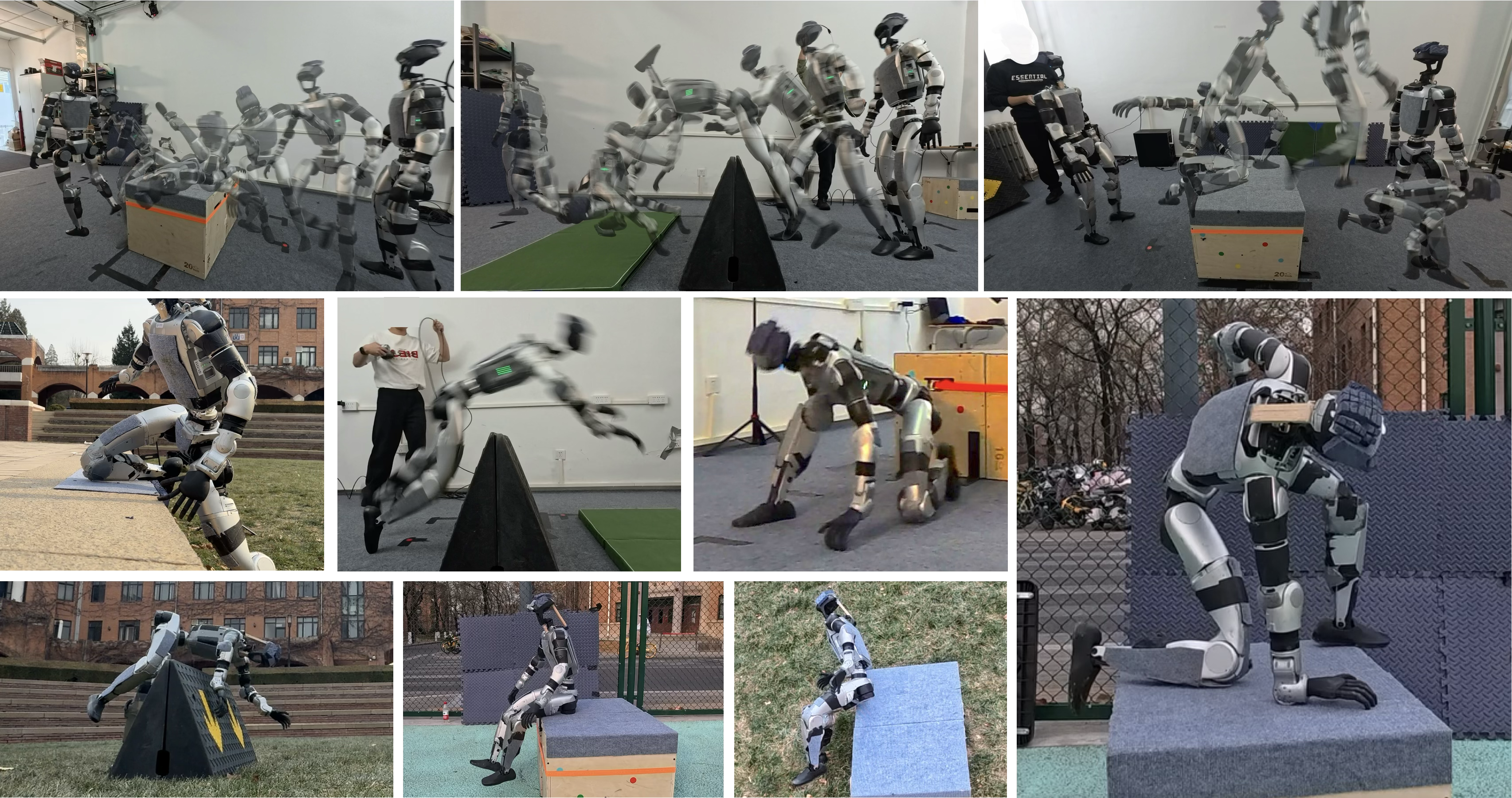}
       \captionof{figure}{\textbf{Deep Whole-Body Parkour.} Our framework enables a humanoid robot to autonomously traverse challenging obstacles that impose strict geometric constraints on robot odometry. The top row visualizes the dynamic execution of vaulting, diving, and jumping using stroboscopic photography. The bottom rows demonstrate the policy's robustness in diverse indoor and outdoor environments. Crucially, the system utilizes onboard perception to proactively adjust its approach trajectory, ensuring precise foot placement and hand contact for successful whole-body interaction.
       \ifanonsubmission
            {} 
        \else
            {Please visit the project website and see the open-sourced infrastructure at https://project-instinct.github.io/deep-whole-body-parkour}
        \fi
       }
       \label{fig:teaser}
       \vspace{-0.1cm}
    \end{center}
    }]
}

\begin{document}

\title{Deep Whole-body Parkour}

\ifanonsubmission
    \author{Author Names Omitted for Anonymous Review. Paper-ID [add your ID here]}
\else
    \author{
    \authorblockN{
        Ziwen Zhuang$^{12}$,
        Shaoting Zhu$^{12}$,
        Mengjie Zhao$^{1}$ and 
        Hang Zhao$^{12}$\authorrefmark{2}
    }
    \authorblockA{$^1$IIIS, Tsinghua University, $^2$Shanghai Qi Zhi Institute}
    \authorblockA{\authorrefmark{2}Corresponding Author}
}
\fi


%

\maketitle

\begin{abstract}
Current approaches to humanoid control generally fall into two paradigms: perceptive locomotion, which handles terrain well but is limited to pedal gaits, and general motion tracking, which reproduces complex skills but ignores environmental capabilities. This work unites these paradigms to achieve perceptive general motion control. We present a framework where exteroceptive sensing is integrated into whole-body motion tracking, permitting a humanoid to perform highly dynamic, non-locomotion tasks on uneven terrain. By training a single policy to perform multiple distinct motions across varied terrestrial features, we demonstrate the non-trivial benefit of integrating perception into the control loop. Our results show that this framework enables robust, highly dynamic multi-contact motions—such as vaulting and dive-rolling—on unstructured terrain, significantly expanding the robot's traversability beyond simple walking or running.
\end{abstract}

\IEEEpeerreviewmaketitle

\section{Introduction}
\begin{figure*}[t]
    \centering
    \includegraphics[width=0.96\linewidth]{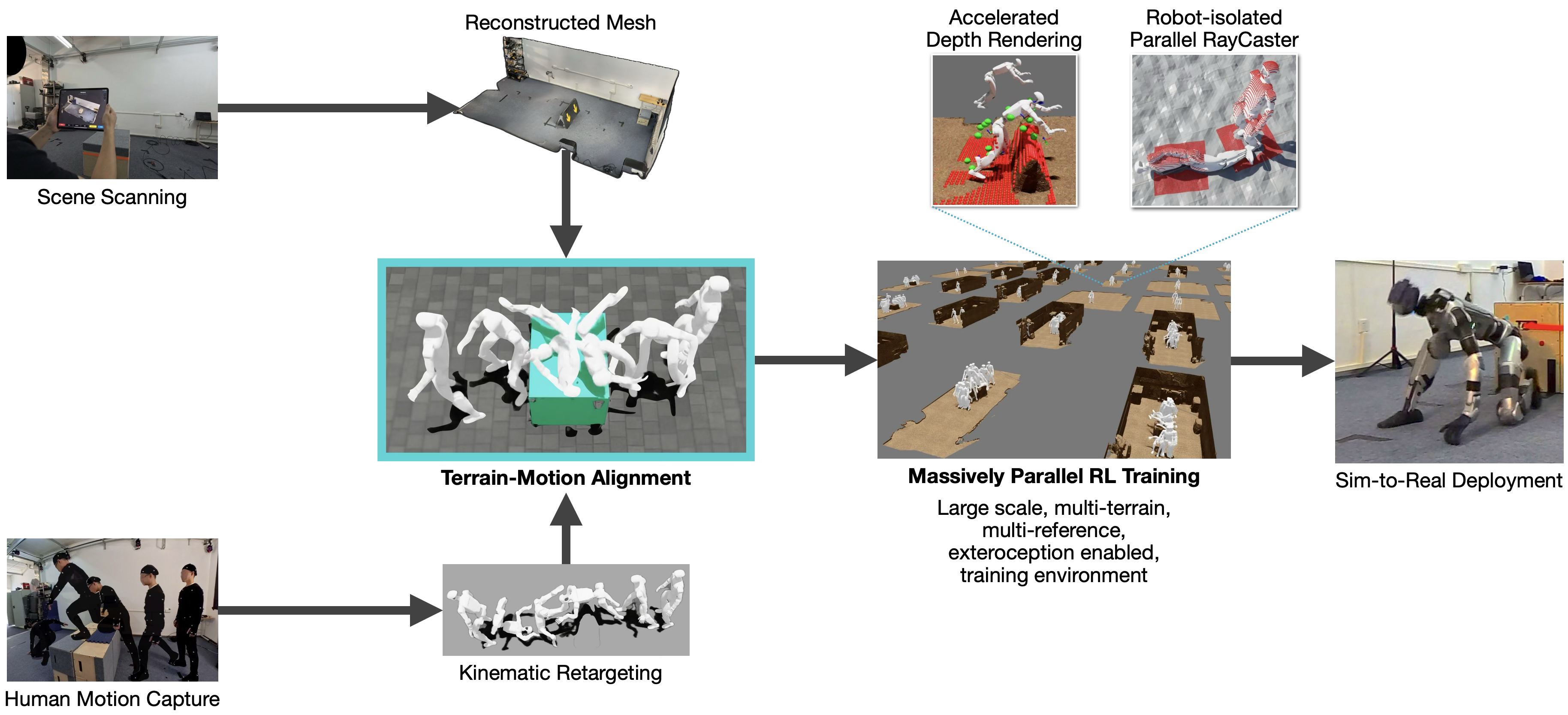}
    \caption{Data-driven whole-body control framework. Real-world environment scans and human demonstrations are processed and aligned to generate feasible motion-terrain pairs. A policy is trained via large-scale reinforcement learning with exteroceptive observations, enabling the robot to replicate agile behaviors in the real world.}
    \label{fig:pipeline}
\end{figure*}

Deep Reinforcement Learning (DRL) has fundamentally transformed the landscape of legged robotics, enabling systems to traverse complex environments with unprecedented robustness. By fusing proprioceptive data with exteroceptive observations—such as depth images or height maps—robots can now adapt their gait patterns to unstructured terrain in real-time~\cite{agarwal_legged_2023, zhang_learning_2024, duan_sim--real_2022}. This paradigm has rapidly evolved from low-speed navigation on static terrain to high-speed, agile maneuvers. Recent advancements have demonstrated that quadrupeds and humanoids are capable of dynamic parkour, autonomously leaping over gaps and climbing large obstacles~\cite{zhuang_robot_2023, zhuang_humanoid_2024, long_learning_2024}. However, despite these aggressive motions, this paradigm remains functionally limited to pedal locomotion. In these frameworks, the robot interacts with the environment exclusively through foot contacts, utilizing the upper body solely for balance rather than for contact-rich interaction. Consequently, while these robots can jump, they lack the capability to execute multi-contact skills—such as vaulting, scrambling, or hand-assisted climbing—that are essential for human-level traversability in constrained spaces.

Conversely, the second paradigm focuses on general motion tracking, a field rooted in computer graphics that prioritizes kinematic  over environmental adaptability. In simulation, previous works such as DeepMimic~\cite{peng_deepmimic_2018} and AMP~\cite{peng_amp_2021} demonstrated physically simulated avatars capable of mastering a rich repertoire of skills—including backflips, martial arts, and parkour vaults—by imitating reference motion data. Recently, this paradigm has successfully bridged the sim-to-real gap, with various studies deploying imitation-based policies on physical humanoids to reproduce expressive behaviors in the real world~\cite{liao_beyondmimic_2025, zhuang_embrace_2025, xue_unified_2025, chen_gmt_2025}. However, a critical limitation persists: these approaches are predominantly environment-agnostic. They rely solely on tracking a pre-recorded trajectory rather than reacting to the geometry of the world. Consider the specific challenge of vaulting onto a high box platform: a blind tracking policy might perfectly reproduce the kinematics of the vault in open space, but without visual feedback, it cannot adjust its jump timing or hand placement to the specific height and distance of the box. As a result, the robot is destined to either collide with the platform or miss the critical hand contacts entirely.

In this work, we propose to bridge these distinct paradigms by integrating exteroceptive depth perception directly into a whole-body motion tracking framework. We introduce a system where the robot learns not merely to mimic a reference animation but to adapt it based on the visual occupancy of the environment. Crucially, the integration of depth perception yields non-trivial robustness to initial conditions. Unlike blind tracking baselines—which require the robot to be placed at a precise, predefined distance from the obstacle to align the playback trajectory—our visually guided policy operates in a closed loop. This allows the system to tolerate significant variance in the setup: for example, if the robot is initialized at varying distances or angles relative to the platform, the policy leverages visual feedback to autonomously adjust its approach gait—shortening or lengthening its steps—to ensure accurate hand placement and successful vaulting. This capability effectively transforms fragile, fixed-trajectory tracking into robust, spatially-aware capabilities suitable for real-world deployment.

\section{Related Works}

\paragraph{Deep Reinforcement Learning for Legged Control}
Deep RL has established a robust standard for locomotion by leveraging proprioceptive history and privileged simulation states~\cite{tan_sim--real_2018, kumar_rma_2021, lee_learning_2020, yang_data_2019}. Through teacher-student frameworks or asymmetric actor-critics, policies learn to estimate terrain properties implicitly~\cite{hwangbo_learning_2019, lee_learning_2020, kumar_rma_2021, nahrendra_dreamwaq_2023}. While highly stable on continuous irregularities, these "blind" agents are methodologically limited to reactive reflexes; without exteroception, they cannot anticipate or plan for discrete obstacles.

\paragraph{Perceptive Locomotion}
Integrating exteroception (e.g., depth or elevation maps) enables agents to modulate foot placement for upcoming terrain, facilitating dynamic parkour behaviors~\cite{miki_learning_2022, agarwal_legged_2022, zhuang_robot_2023, zhuang_humanoid_2024}. However, these systems typically rely on low-dimensional velocity commands ($v_x, v_y, \omega_z$), leading to a \textit{task under-specification} problem for humanoids. This low-bandwidth interface cannot disambiguate complex interaction modes—such as vaulting versus jumping—thereby restricting the robot to simple pedal locomotion despite its expressive morphology.

\paragraph{Data-Driven Humanoid Motion Tracking}
Motion tracking approaches, such as DeepMimic~\cite{peng_deepmimic_2018} and AMP~\cite{peng_amp_2021}, utilize reference motions as a dense task specification, successfully transferring agile skills to physical hardware~\cite{luo_perpetual_2023, xue_unified_2025, he_learning_2024, he_hover_2024, liao_beyondmimic_2025}. However, these frameworks fundamentally operate under a \textit{planar environment assumption}. They treat locomotion purely as kinematic reproduction on flat ground, disregarding the necessity to perceive terrain-dependent information. Consequently, without geometric awareness, they lack the capability to traverse non-planar structures or adapt the reference motion to physical obstacles.

\section{Method}

\subsection{Dataset Curation and Environment Generation}
\label{subsec:dataset}

\paragraph{Motivation and Capture}
While large-scale motion datasets such as AMASS~\cite{mahmood_amass_2019} provide extensive human kinematic data, and OMOMO~\cite{li_object_2023} introduces object manipulation, they generally lack dynamic, whole-body interactions with large-scale geometry. Agile parkour maneuvers—such as vaulting, climbing, and hurdling—rely critically on precise contacts between the agent and the terrain. To bridge this gap, we curate a custom dataset that strictly couples human dynamics with accurate environmental geometry.

We employ an optical motion capture system to record expert human actors performing parkour maneuvers on physical obstacles. To ensure high-fidelity spatial correspondence between the motion and the terrain, we simultaneously digitize the physical scene using a LiDAR-enabled iPad Pro (via the 3D Scanner App). This process yields a reconstructed mesh that is spatially aligned with the captured motion trajectories (Figure~\ref{fig:pipeline}).

\paragraph{Motion Retargeting}
The raw human motion is retargeted to the Unitree G1 humanoid robot using the GMR framework~\cite{araujo_retargeting_2025}. This process utilizes optimization-based kinematic filtering, followed by manual keyframe adjustment, to ensure the resulting trajectories are physically feasible for the robot's morphology. Special attention is paid to enforcing valid contact constraints and eliminating high-frequency capture noise.

\paragraph{Procedural Environment Generation}
To ensure the policy generalizes to diverse environments rather than overfitting to the specific spatial constraints of our capture laboratory, we post-process the scanned meshes for simulation. We isolate the functional geometry (e.g., obstacles, platforms, and rails) by segmenting out surrounding walls, ceilings, and extraneous laboratory context. This yields a set of canonical, context-agnostic obstacle meshes.

Finally, we integrate these assets into NVIDIA Isaac Lab to create a massive-parallel training environment. We treat the retargeted motion and its corresponding obstacle mesh as a single \textit{paired instance}. These pairs are procedurally instantiated across the simulation grid, creating an open-field training setup shown in Figure~\ref{fig:pipeline}. By stripping the scene of collision group artifacts (like room boundaries) and randomizing the placement of these motion-terrain pairs, we ensure the agent learns to condition its behavior strictly on local obstacle geometry.

\paragraph{Massively Parallel Ray-Caster for Isolated Multi-Agent Training}
To train a perceptive whole-body control policy within a unified framework, high-throughput depth simulation across thousands of parallel environments is essential. While IsaacLab provides GPU-accelerated simulation, clear limitations exist regarding sensor rendering: standard implementations struggle to simultaneously render complex moving articulations and static terrain while strictly isolating parallel environments. Specifically, distinct robots must not perceive "ghost" instances of other robots residing in different environments within the same physics scene.

To address this, we implement a custom, highly optimized ray-caster utilizing Nvidia Warp. To maximize memory efficiency, we employ mesh instancing; we collect a set of collision mesh prototypes aligned with critical articulations and maintain a global batch of transform matrices for all active instances. We introduce a collision grouping mechanism where static terrain is assigned a universal group ID ($-1$), which is visible to all agents. Conversely, each robot is assigned a unique collision group ID to ensure disjoint perception.

\begin{algorithm}[t]
\caption{Massively Parallel Grouped Ray-Casting}
\label{alg:ray_casting}
\begin{algorithmic}[1]
\Require 
    Ray inputs $\mathbf{R} = \{(\mathbf{o}_i, \mathbf{d}_i, g_i)\}_{i=1}^{N_{rays}}$ (origin, direction, group ID), 
    Mesh instances $\mathcal{M}$, 
    Transforms $\mathbf{T}$, 
    Mesh Group IDs $\mathcal{G}_{mesh}$
\Ensure Depth map $\mathbf{D} \in \mathbb{R}^{N_{rays}}$

\Statex \textbf{\textit{Phase 1: Acceleration Structure Construction (Pre-compute)}}
\State Initialize hash map $\mathcal{H}_{map}: \text{Group ID} \rightarrow \text{List of Mesh Indices}$
\For{each mesh instance $m_j \in \mathcal{M}$}
    \State $gid \leftarrow \mathcal{G}_{mesh}[j]$
    \State Append $j$ to $\mathcal{H}_{map}[gid]$
\EndFor
\State Define global static group $G_{static} \leftarrow -1$

\Statex \textbf{\textit{Phase 2: Parallel Rendering (GPU Kernel)}}
\For{each ray $r_i \in \mathbf{R}$ in parallel}
    \State $t_{min} \leftarrow \infty$
    \State $g_{agent} \leftarrow r_i.g_i$
    
    \State \Comment{1. Check Global Static Terrain}
    \For{each mesh index $idx$ in $\mathcal{H}_{map}[G_{static}]$}
        \State $t \leftarrow \Call{Intersect}{r_i, \mathcal{M}[idx], \mathbf{T}[idx]}$
        \If{$t < t_{min}$} $t_{min} \leftarrow t$ \EndIf
    \EndFor
    
    \State \Comment{2. Check Agent-Specific Dynamic Objects (Skip others)}
    \For{each mesh index $idx$ in $\mathcal{H}_{map}[g_{agent}]$}
        \State $t \leftarrow \Call{Intersect}{r_i, \mathcal{M}[idx], \mathbf{T}[idx]}$
        \If{$t < t_{min}$} $t_{min} \leftarrow t$ \EndIf
    \EndFor
    
    \State $\mathbf{D}[i] \leftarrow t_{min}$
\EndFor
\State \Return $\mathbf{D}$
\end{algorithmic}
\end{algorithm}

A naive implementation would require every cast ray to iterate through the global table of mesh group IDs to determine visibility, resulting in a complexity linear to the total number of meshes in the scene $O(N)$. With tens of thousands of articulation parts in massively parallel training, this becomes a severe bottleneck. We propose a \textit{Precomputed Grouped Ray-Casting} strategy to accelerate this process, as described in Algorithm~\ref{alg:ray_casting}. We pre-compute a mapping from collision group indices to mesh instance IDs. Consequently, during the ray-marching phase, a ray associated with a specific agent iterates solely over the global static meshes and the specific subset of dynamic meshes belonging to that agent's group. This reduces the search space significantly, resulting in a $10\times $ increase in rendering speed compared to the naive baseline.

\subsection{Training settings}
We train the policy using both proprioception and exteroception, specifically through depth image. Following BeyondMimic~\cite{liao_beyondmimic_2025} and VMP~\cite{serifi_vmp_2024}, we design the reward terms by tracking the robot local poses in the relative frame and the global pose of the robot root link.

\paragraph{Relative Frame} By defining the relative frame $T_\text{rel}$, we define the reference root transform $T_\text{ref}$ as the transform of the root link in the motion reference, and the robot actual root transform $T_\text{robot}$. The relative frame $T_\text{rel}$ is in between. $T_\text{rel}$ has the same x-y coordinate as $T_\text{robot}$ but the same z coordinate as $T_\text{ref}$. $T_\text{rel}$ has the same roll-pitch angle as $T_\text{ref}$ but the same yaw angle as $T_\text{robot}$.

\begin{figure}[h!]
    \centering
    \includegraphics[width=0.9\linewidth]{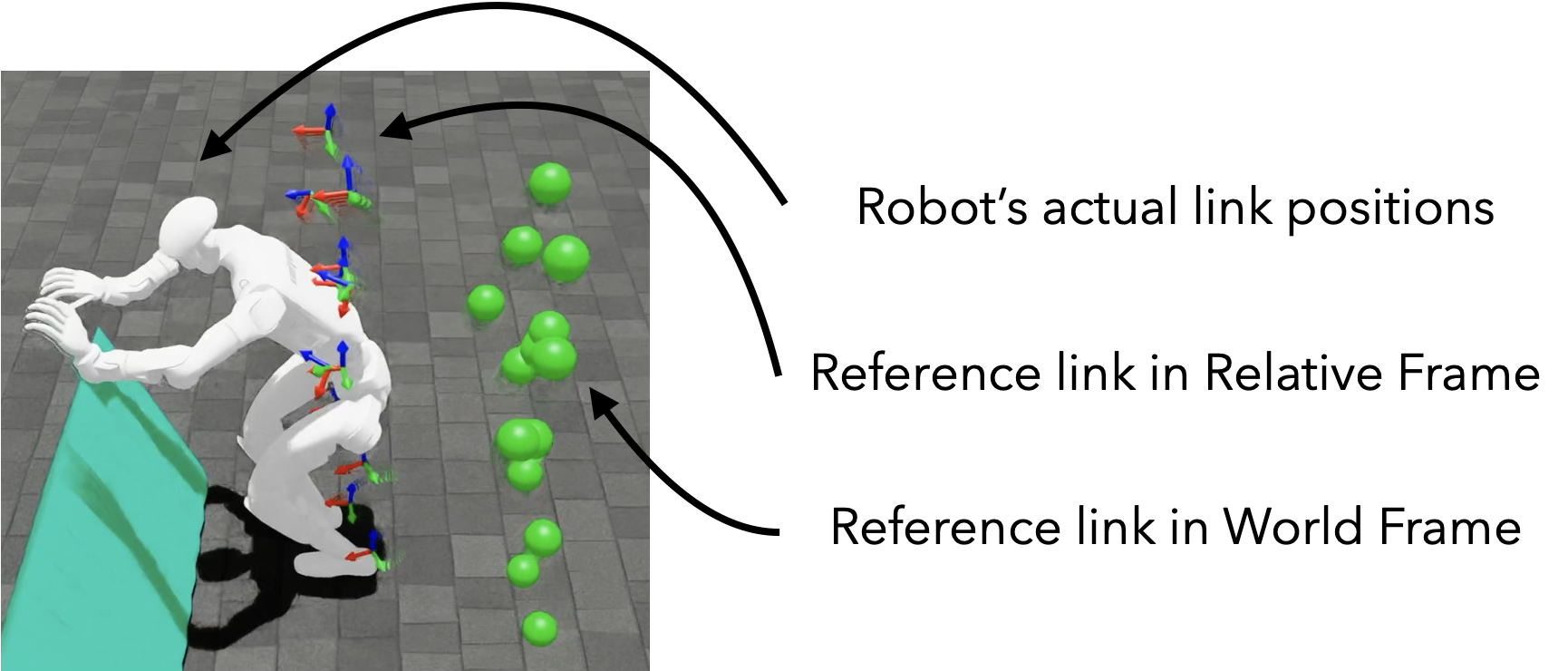}
    \caption{We illustrate the basic concept of relative frame. It has the same x-y and yaw coordinates as the robot base frame and has the same z, roll and pitch coordinate in the reference frame.}
    \label{fig:relative-frame}
\end{figure}

Mathematically, we define $T_\text{rel}$ as a $4 \times 4$ matrix where
$$
T_{\text{rel}} = \begin{bmatrix} R_{\text{rel}} & \mathbf{p}_{\text{rel}} \\ \mathbf{0} & 1 \end{bmatrix}
$$
where the translation $\mathbf{p}_{\text{rel}}$ combines the planar position of the robot with the vertical position of the reference,
$$
\mathbf{p}_{\text{rel}} = \begin{bmatrix} x_{\text{robot}} \\ y_{\text{robot}} \\ z_{\text{ref}} \end{bmatrix}
$$
and the rotation $R_{\text{rel}}$ combines the heading (yaw) of the robot with the tilt (roll and pitch) of the reference
$$
R_{\text{rel}} = R_z(\psi_{\text{robot}}) \, R_y(\theta_{\text{ref}}) \, R_x(\phi_{\text{ref}})
$$
More intuitive illustration is shown in Figure~\ref{fig:relative-frame}.

\paragraph{Rewards Definition} We adopt the policy training using asymmetric PPO. We feed the policy with the synchronized future joint position reference, future joint velocity reference, future root position reference (in reference transform) and future robot rotation reference (in actual robot root transform). Then we feed the policy with one frame of noised depth image and a historical proprioception observation with 8 frames. For critic network, we feed the network with the synchronized future joint position reference, future joint velocity reference, future root position reference (in the actual robot root transform), future robot rotation reference (in the actual robot root transform), actual robot link transforms (in robot's root transform), height-scan to the terrain, and a historical proprioception observation with 8 frames.

Following BeyondMimic~\cite{liao_beyondmimic_2025}, we train with simplified tracking rewards as follows:
\begin{itemize}
    \item Global root position reward based on distance to the reference position.
    \item Global root rotation reward based on rotation difference to the reference rotation in axis angle.
    \item Local link position reward based on key links position differences in relative frame.
    \item Local link rotation reward based on key links rotation differences in relative frame in the magnitude of axis angle.
    \item Global link linear velocity reward based on the key links linear velocity difference in the world frame.
    \item Global link angular velocity reward based on the key links angular velocity difference in the world frame.
    \item Action rate penalty.
    \item Joint position limit penalty.
    \item Undesired contact penalty.
    \item Applied torque limits penalty.
\end{itemize}

\paragraph{Adaptive Sampling}
\begin{algorithm}[t]
\caption{Adaptive Sampling with Failure-Based Curriculum}
\label{alg:adaptive_sampling}
\begin{algorithmic}[1]
\Require Set of reference motions $\mathcal{M} = \{m_1, \dots, m_K\}$
\Require Bin duration $\Delta t_{max} = 1.0s$
\Require Smoothing kernel $\mathcal{K}$ (e.g., Exponential)
\State \textbf{Initialize:} Discretize each $m_k$ into bins $B_{k} = \{b_{k,0}, \dots, b_{k, T_k}\}$
\State \textbf{Initialize:} Failure counts $F_{k,t} \leftarrow 0$ and Sampling Weights $W_{k,t} \leftarrow \text{Uniform}$

\For{each training iteration}
    \State \Comment{\textit{1. Adaptive Selection}}
    \State Normalize weights: $P(k, t) \leftarrow \frac{W_{k,t}}{\sum_{k',t'} W_{k',t'}}$
    \State Sample motion index $k$ and start time $t_{start} \sim P(k, t)$
    
    \State \Comment{\textit{2. Policy Rollout}}
    \State Initialize robot state at $S_{t_{start}}$ from motion $m_k$
    \State Execute policy $\pi$ until completion or failure at $t_{fail}$
    
    \If{Failure occurred at $t_{fail}$}
        \State Identify discrete bin index $\hat{t}$ corresponding to $t_{fail}$
        \State $F_{k, \hat{t}} \leftarrow F_{k, \hat{t}} + 1$ \Comment{Increment raw failure count}
        
        \State \Comment{\textit{3. Update Sampling Distribution}}
        \For{each bin $j$ in motion $m_k$}
            \State Apply smoothing: $\tilde{F}_{k,j} \leftarrow \sum_{i} F_{k,i} \cdot \mathcal{K}(j, i)$
        \EndFor
        \State Update weights: $W_{k,t} \leftarrow \tilde{F}_{k,t} + \epsilon$ \Comment{Add $\epsilon$ to ensure non-zero probability}
    \EndIf
\EndFor

\end{algorithmic}
\end{algorithm}
We formulate the training environment as a Markov Decision Process (MDP) and employ an adaptive sampling strategy to facilitate curriculum learning. To handle variable-length reference trajectories, we discretize each motion into a sequence of temporal bins with a maximum duration of $t_\text{bin}=1.0s$. This standardization allows us to treat distinct motion segments uniformly, regardless of the total trajectory length. During training rollouts, we track the agent's performance; if a termination or failure occurs, we increment the failure counter for the bin corresponding to the failure timestep. To maintain training stability and prevent abrupt distribution shifts, we apply temporal smoothing to these raw failure counts. This operation converts discrete integer counts into continuous failure scores, effectively diffusing the difficulty signal to adjacent timesteps. Finally, we utilize these smoothed scores to construct a probability distribution over all bins across all motions. The training reset state (which motion and what start time) is then sampled based on these weights, ensuring the agent prioritizes high-failure scenarios. The detailed implementation logic is shown in Algorithm~\ref{alg:adaptive_sampling}.

\paragraph{Early Timeout with Stuck Detection} However, the randomization terms are not fully applicable to uniform randomization across the entire motion sequence. For example, the uniform randomization in sampling the robot state from the reference trajectory will place the robot in the middle of some terrain. This generates unsolvable situations for the policy to learn from. To resolve this issue, we truncate the trajectories that have caused the robot to get stuck over a given time limit at the start of the episode.

\paragraph{Network inputs} Since we design the training framework in one stage, the policy network does not have any odometry-related information. While based on the principle of asymmetric actor critic design, the critic network will have odometry-related information for more accurate value function prediction. Referring to \citet{zhuang_embrace_2025, chen_gmt_2025, ze_twist_2025}, we train both policy network and critic network with 10 frames of future reference frame. Each frame has 0.1s interval with respect to the previous frame, which sums up to 1.0s future motion expectation.

For policy network, the inputs contain:
\begin{itemize}
    \item Future joint position reference sequence.
    \item Future joint velocity reference sequence.
    \item Future base position reference sequence with respect to the base position of the reference frame.
    \item Future base rotation reference sequence with respect to the base rotation of the actual robot in simulator.
    \item Single frame of noised depth image from the robot's camera.
    \item 8 frames of historical proprioception.
\end{itemize}

For critic network, the inputs contain:
\begin{itemize}
    \item Future joint position reference sequence.
    \item Future joint velocity reference sequence.
    \item Future base position reference sequence with respect to the base position of the actual robot in simulator.
    \item Actual key links position in robot's base frame.
    \item Actual key links rotation (tan-norm) in robot's base frame.
    \item Height scan data.
    \item 8 frames of historical proprioception.
\end{itemize}

\paragraph{Network architecture} Since the motion tracking task is only spread across a handful number of motions, we use a straight-forward design, with a CNN encoder encoding the depth image and feed to a 3 layer MLP network together with the proprioception history. The detailed parameters are listed in the appendix.

\subsection{Bridging sim-to-real gap in depth perception}
\begin{figure*}[t!]
    \centering
    \includegraphics[width=0.8\linewidth]{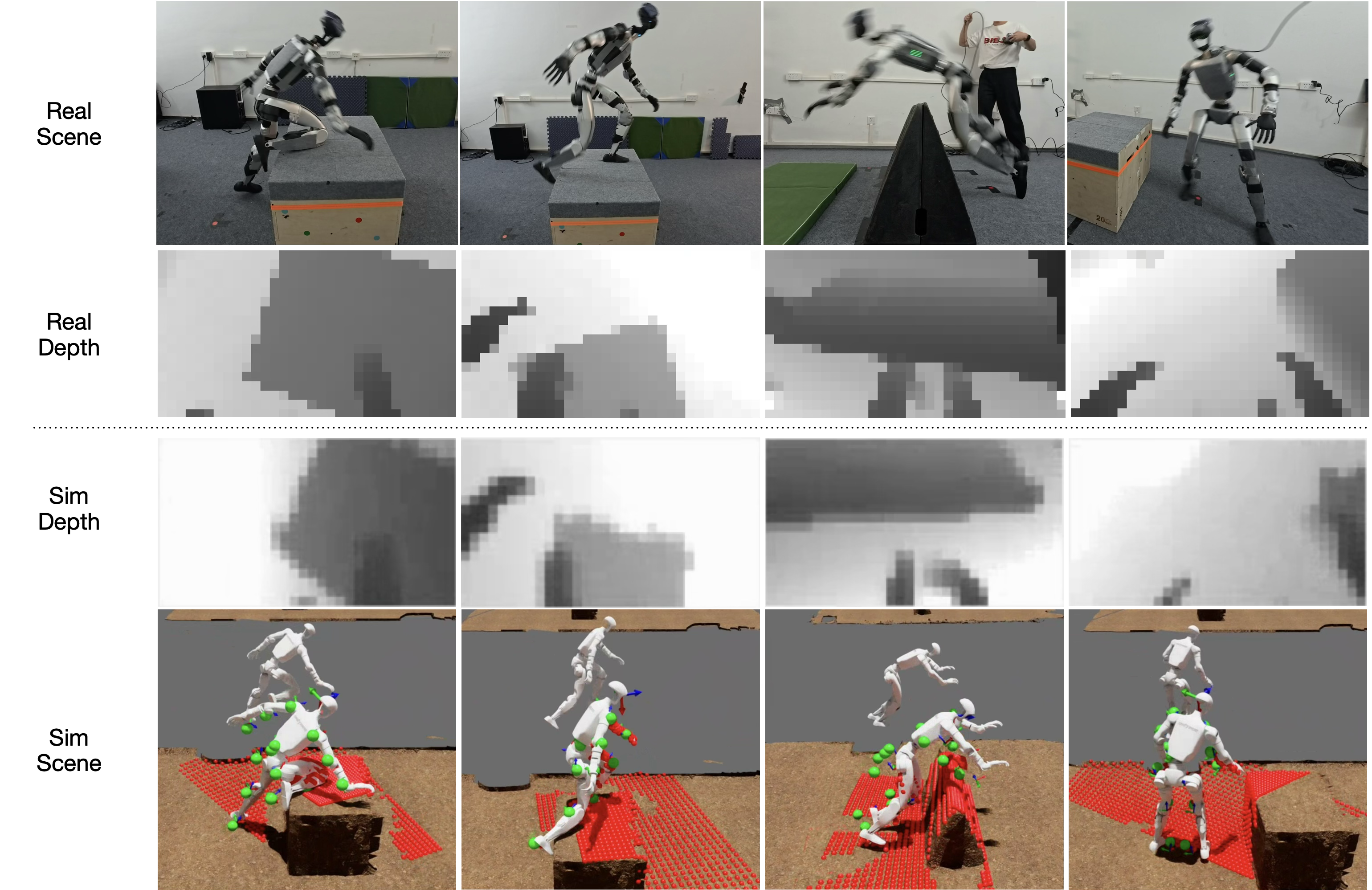}
    \caption{To bridge the sim-to-real gap in depth visualization, we apply several noise patterns in simulation and applied inpainting algorithm from GPU-based OpenCV implementation.}
    \label{fig:sim-to-real-depth}
\end{figure*}
It should be noted that we train our policy using synchronized depth image observations. We deploy the policy using 50Hz depth image using RealSense. However, the builtin filtering algorithm from pyrealsense does not meet the 50Hz requirement on the Nvidia Jetson's CPU. We use the inpainting algorithm from a GPU-based OpenCV implementation. Due to occasionally artifacts pattern in the real world because of reflection, motion blur or stereo algorithm error, we add gaussian noise, patched artifacts to the depth image in simulation similar to \citet{zhuang_robot_2023, zhuang_humanoid_2024}. As shown in Figure~\ref{fig:sim-to-real-depth}, the depth images fed into the network between simulation and real-world are quite similar in terms of pixel pattern.

\subsection{Deployment}
To deploy the policy onboard with depth sensing running at 50Hz, we adopt ONNX as our neural network accelerator. Even though the motions are trained to a single depth-based neural network, we still need to choose the motion reference as input. Thus, we implemented a state machine mechanism to select the motion during deployment.

Also, to generate running logs during the real-world experiments while reducing the impact on the neural network process as low as possible, we run rosbag recording in a separate process. Also, we bind the CPUs separately for running the neural network and rosbag recording process to prevent further CPU scheduling.

Considering the policy is trained with root position reference in motion reference frame, this part of the observation does not depend on the odometry of the robot, neither in simulation nor in the real-world. Thus, we do not need any odometry system providing global position error between the robot and the motion reference. We provide the exact same joint position, joint velocity reference to the policy as it is trained. We provide the relative rotation difference based on the real-time IMU readings of the robot, before which the start heading of the motion reference was aligned to the robot's actual heading. After correcting the heading of the motion reference, as well as the direction of the reference position sequence, we provide the policy with only the position reference relative to the frame of reference position without any odometry information.

\section{Experiments}
In this section, we aim to raise and answer some questions, addressing the pros and cons of adding exteroception to the highly agile motion tracking system. Also, to test the generalization across multiple terrains and multiple motions in the same terrain, we curate 4 motions across 3 different terrains. One is the triangular barrier formed by two road ramps. One is a big wooden box the size of $0.5m \times 0.6m \times 0.4m$. The last terrain is only the flat terrain, as the complementary experiment to show that the system can still perform non-interactive motions in flat terrain. The 4 motions consist of `kneel climb', `roll vault', `dive roll', and `jump sit'.

We first run real-world experiments to the effectiveness of our pipeline that can be successfully deployed in the real robot, then we analyze the effect of adding depth vision to the system by proposing the following questions:
\begin{enumerate}
    \item What benefit does the additional depth vision provide to this agile whole-body control system that interacts with the unstructured terrain?
    \item Are there any drawbacks when depth vision is provided as additional information? Is it robust to unseen but trivial distractors in the scene?
    \item Throughout the entire training pipeline, what contributes the most to the successful sim-to-real deployment on the real robot without any odometry system?
\end{enumerate}

\subsection{Real world experiment}

In the real world experiment, we deploy our entire inference system using ROS2 on a Unitree G1 29-DOF, September 2025 version. The neural network is accelerated using ONNX (CPU version) onboard. We acquire the depth image using an Intel RealSense D435i installed on the head of the Unitree G1 in an individual process separate from the network inference process. We integrate a simple walking process to keep balance when the perceptive parkour policy is not activated. At each test, we control the robot to walk to a rough starting point with no accurate odometry or localization system. We then trigger the motion tracking policy with a specific motion reference trajectory. The system switches back to a walking policy when the motion reference trajectory is exhausted.

We perform real-world tests both indoors and outdoors. With indoor environments, we were able to collect running logs using a network cable from another computer. In outdoor environments, we only connect to the onboard compute without an additional logging system. Please notice that, in outdoor environments, the starting point of the motion is not determined. We only put the robot in front of the obstacle without any further calibration.

\subsection{Non-trivial benefit of depth vision}
\begin{table*}[t]
\centering
\caption{Comparison of training variants across four motions. Each motion is evaluated using two metrics ($\text{MPJPE}_g$ and $\text{MPJPE}_b$).}
\label{tab:mpjpe-ablation}
\begin{tabular}{lcccccccc}
\toprule
\multirow{2}{*}{\textbf{Training Variant}} & \multicolumn{2}{c}{\textbf{Dive Roll}} & \multicolumn{2}{c}{\textbf{Kneel Climb}} & \multicolumn{2}{c}{\textbf{Roll Vault}} & \multicolumn{2}{c}{\textbf{Jump Sit}} \\
\cmidrule(lr){2-3} \cmidrule(lr){4-5} \cmidrule(lr){6-7} \cmidrule(lr){8-9}
& $\text{MPJPE}_g$ & $\text{MPJPE}_b$ & $\text{MPJPE}_g$ & $\text{MPJPE}_b$ & $\text{MPJPE}_g$ & $\text{MPJPE}_b$ & $\text{MPJPE}_g$ & $\text{MPJPE}_b$ \\
\midrule
\multicolumn{9}{l}{\textit{Testing with no position randomization}} \\
\midrule
w/o depth (BeyondMimic) & 0.1365 & 0.0462 & 0.1162 & 0.0378 & 0.1109 & 0.0358 & 0.1909 & 0.0341 \\
w/o stuck detection     & 0.1367 & 0.0464 & 0.1073 & 0.0394 & 0.1162 & 0.0393 & 0.1170 & 0.0389 \\
local frame reward      & 0.1213 & 0.0242 & 0.1145 & 0.0383 & 0.1131 & 0.0332 & 0.1358 & 0.0299 \\
ours                    & 0.1468 & 0.0459 & 0.1063 & 0.0388 & 0.1123 & 0.0351 & 0.1197 & 0.0362 \\
\midrule
\multicolumn{9}{l}{\textit{Testing with position randomization}} \\
\midrule
w/o depth (BeyondMimic) & 0.2267 & 0.0477 & 0.1822 & 0.0405 & 0.1969 & 0.0385 & 0.2693 & 0.0377 \\
w/o stuck detection     & 0.1661 & 0.0456 & 0.1449 & 0.0412 & 0.1461 & 0.0316 & 0.1322 & 0.0354 \\
local frame reward      & 0.1892 & 0.0389 & 0.1477 & 0.0402 & 0.1453 & 0.0311 & 0.1312 & 0.0301 \\
ours                    & 0.1649 & 0.0477 & 0.1218 & 0.0399 & 0.1421 & 0.0371 & 0.1446 & 0.0382 \\
\bottomrule
\end{tabular}
\end{table*}

\begin{figure}
    \centering
    \includegraphics[width=0.9\linewidth]{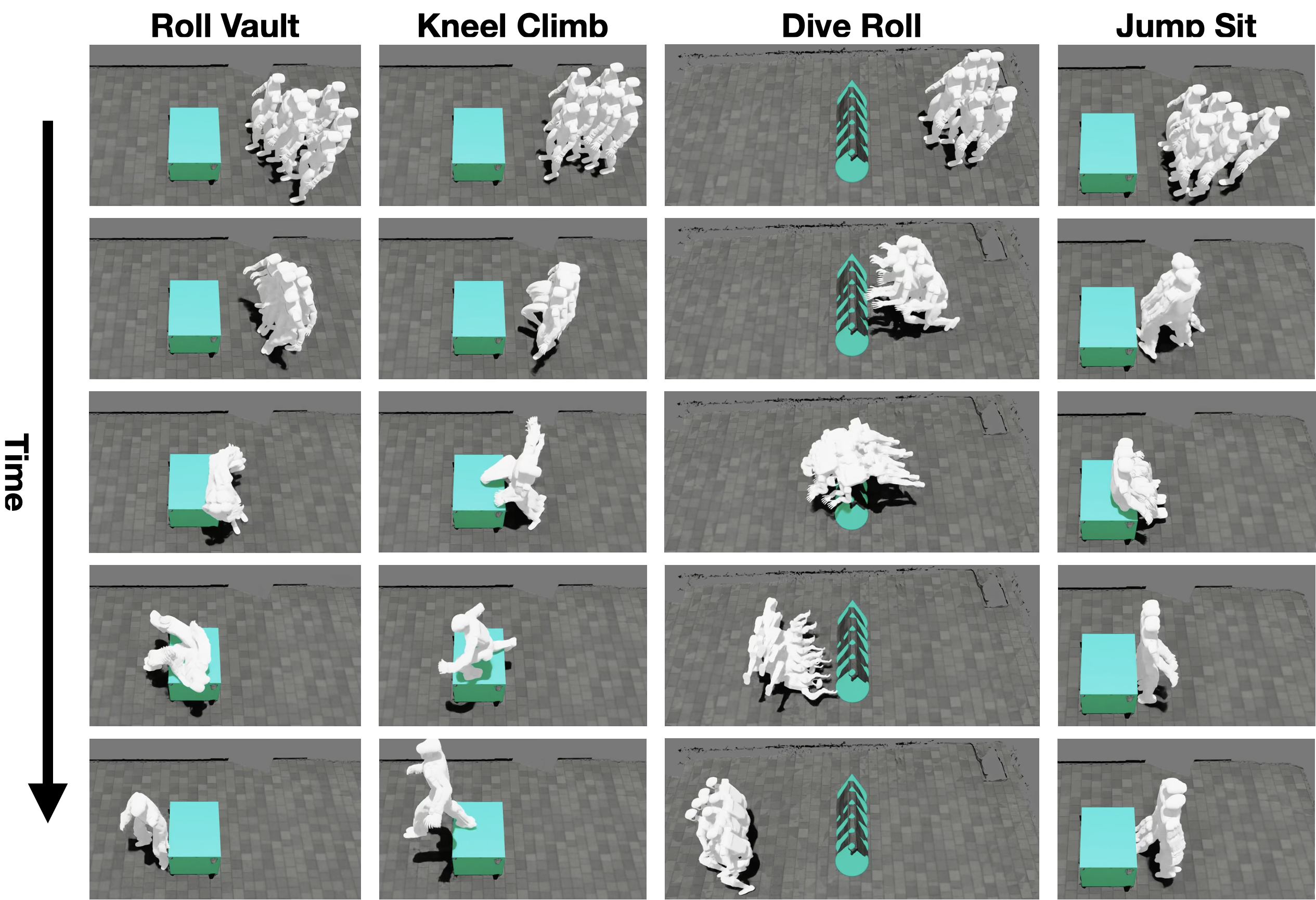}
    \vspace{-6pt}
    \caption{We show a series of direct example, illustrating the positional convergence when introducing depth vision in this end-to-end motion tracking system. As shown in this figure, the position converges automatically even before the scene interaction. Otherwise, for difficult motions like roll-vaulting and kneel-climbing, misplace at the scale of $0.4m$ will lead to catastrophic failure of the scene-interaction task.}
    \label{fig:motion-convergence}
\end{figure}

We investigate how exteroception contributes to the success and generalization of these agile whole-body motions. We uniformly spawn the robot around the initial starting position of the reference motion to test to what extent the policy is able to recover and successfully finish the motion. As shown in Figure~\ref{fig:motion-convergence} we visualize the convergence of the robots in a batch, which start at different positions. As time goes on, all robots converge to the target obstacle and successfully finish the motion tracking task. Even though all trainings randomize the starting position at $0.3m \times 0.3m$ horizontally around the reference motion. We test the motion convergence behavior with the range of $0.5m \times 0.5m$ horizontally around the reference starting point.

\begin{figure}[h!]
    \centering
    \includegraphics[width=1.0\linewidth]{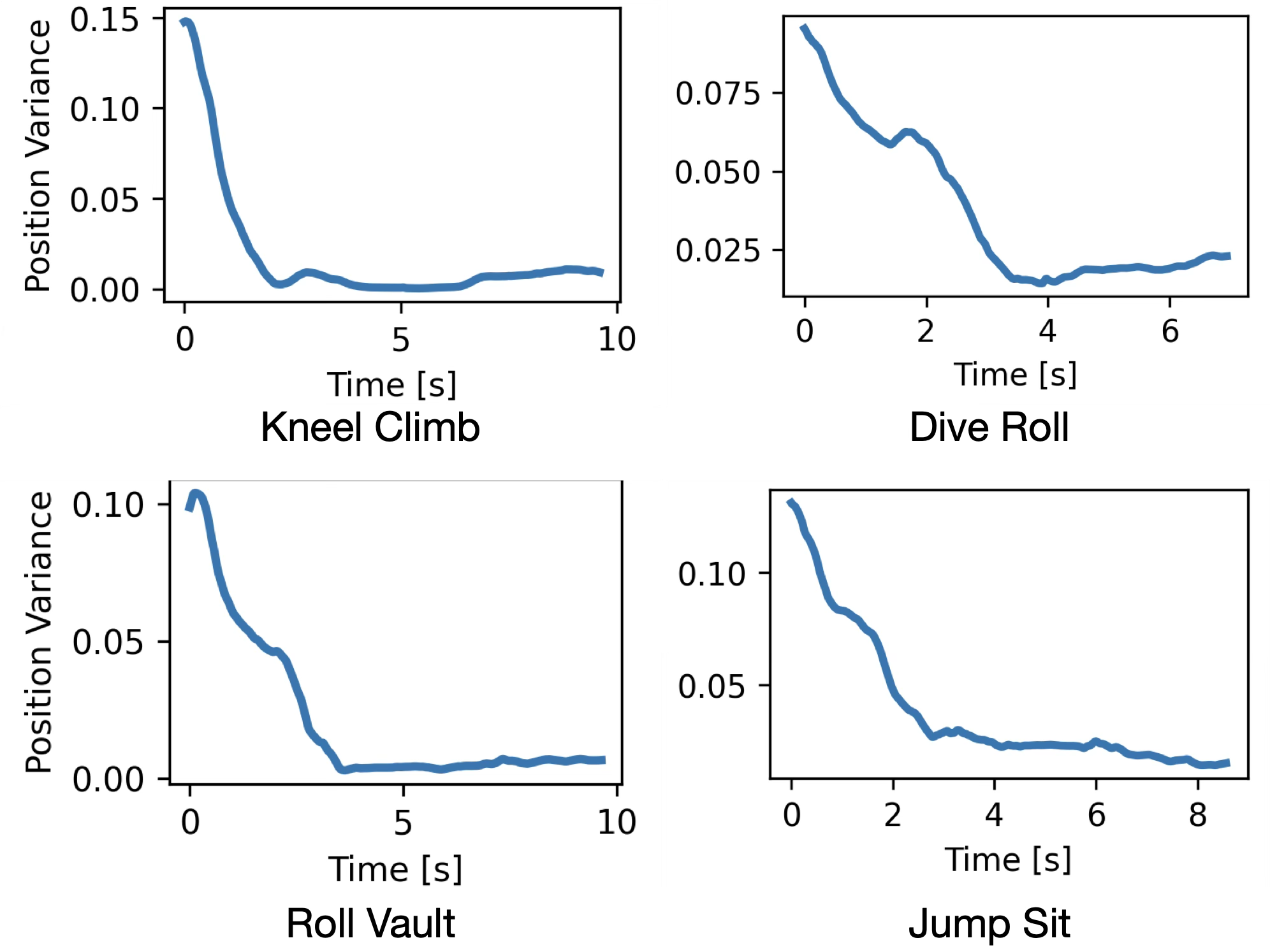}
    \caption{We show the $x-y$ position variance in a single batch of motion reference example to illustrate the emergence of positional correction ability when introducing depth vision to the end-to-end motion tracking system.}
    \label{fig:position-variance}
\end{figure}
We then plot the position variance in x-y coordinate to further verify this observation in Figure~\ref{fig:position-variance}. The position variance drastically converges to a really low value. When the motion reaches its final stage, some of the position variances increase due to the common visual features from the depth camera.

\begin{figure}[h!]
    \centering
    \includegraphics[width=1.0\linewidth]{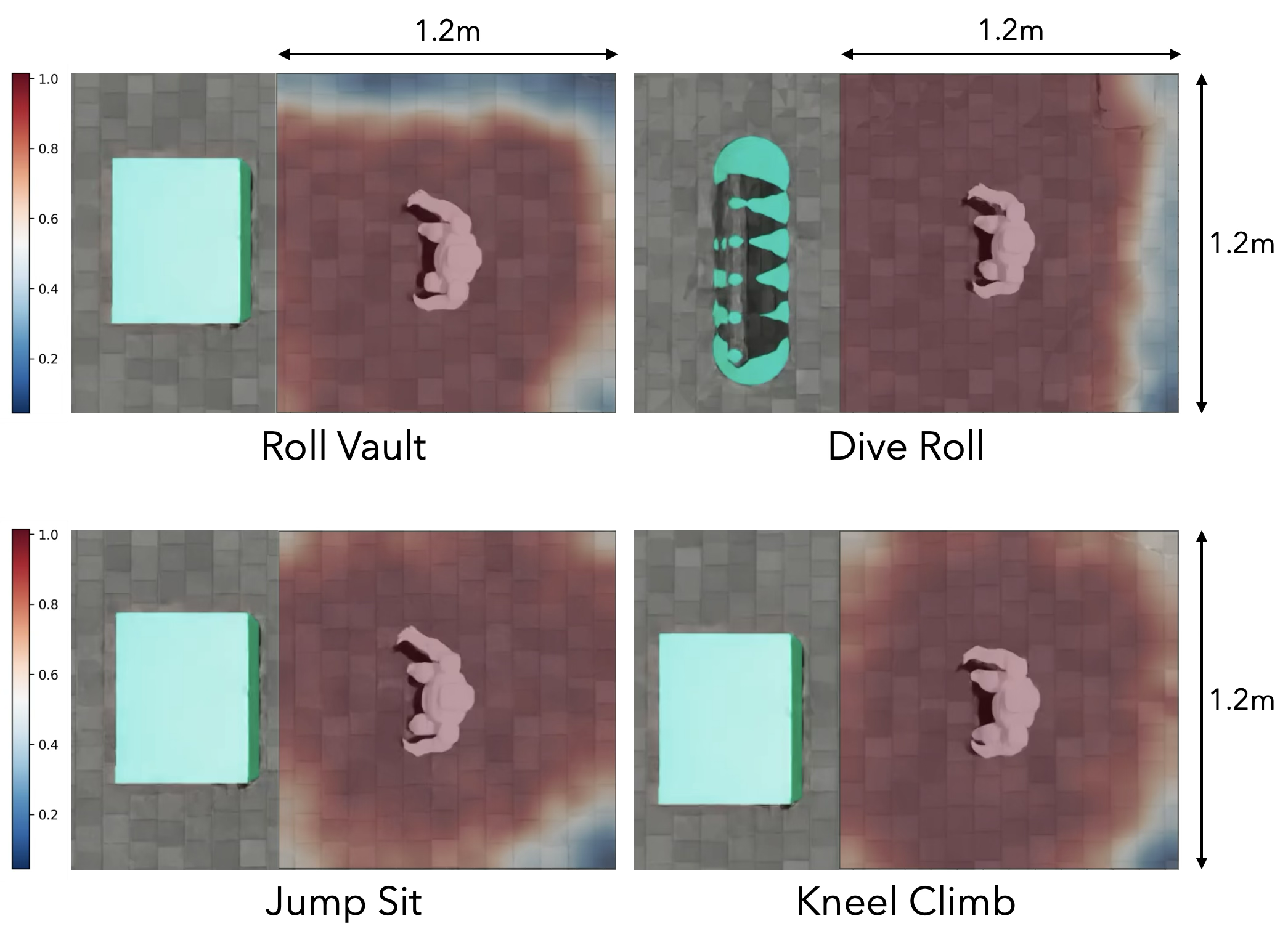}
    \caption{We do a grid search around the starting position of the motion reference frame. We plot the headmap of the success rate of each motion at a $1.2m \times 1.2m$ space. Red suggests the $100\%$ success rate, while dark blue suggests the $0\%$ success rate.}
    \label{fig:success-rate-heatmap}
\end{figure}
To further test the generalization ability of this visual-guided motion tracking pipeline, we perform a more aggressive out-of-distribution tests, even though the randomization during training is only from $-0.15m$ to $+0.15m$ in $x$ and $y$ coordinate. We plot the success rate headmap from $-0.6m$ to $+0.6m$ around the initial motion reference starting point. This forms a $1.2m \times 1.2m$ range of initialization. As shown in Figure~\ref{fig:success-rate-heatmap}, the success rate of the motion tracking system is still $100\%$. It only drops at the boundary of these experiments.

\subsection{Vision robustness}
In this section, we aim to study the robustness of this additional depth information in the motion tracking system. We build several scenes in the simulator that have never been used during training. They are the out-of-distribution cases for the trained motion tracking network. Considering the reference motions are still the same, the added entities in the scene should block the robot's dynamics as little as possible.

\begin{figure}[]
    \centering
    \includegraphics[width=0.9\linewidth]{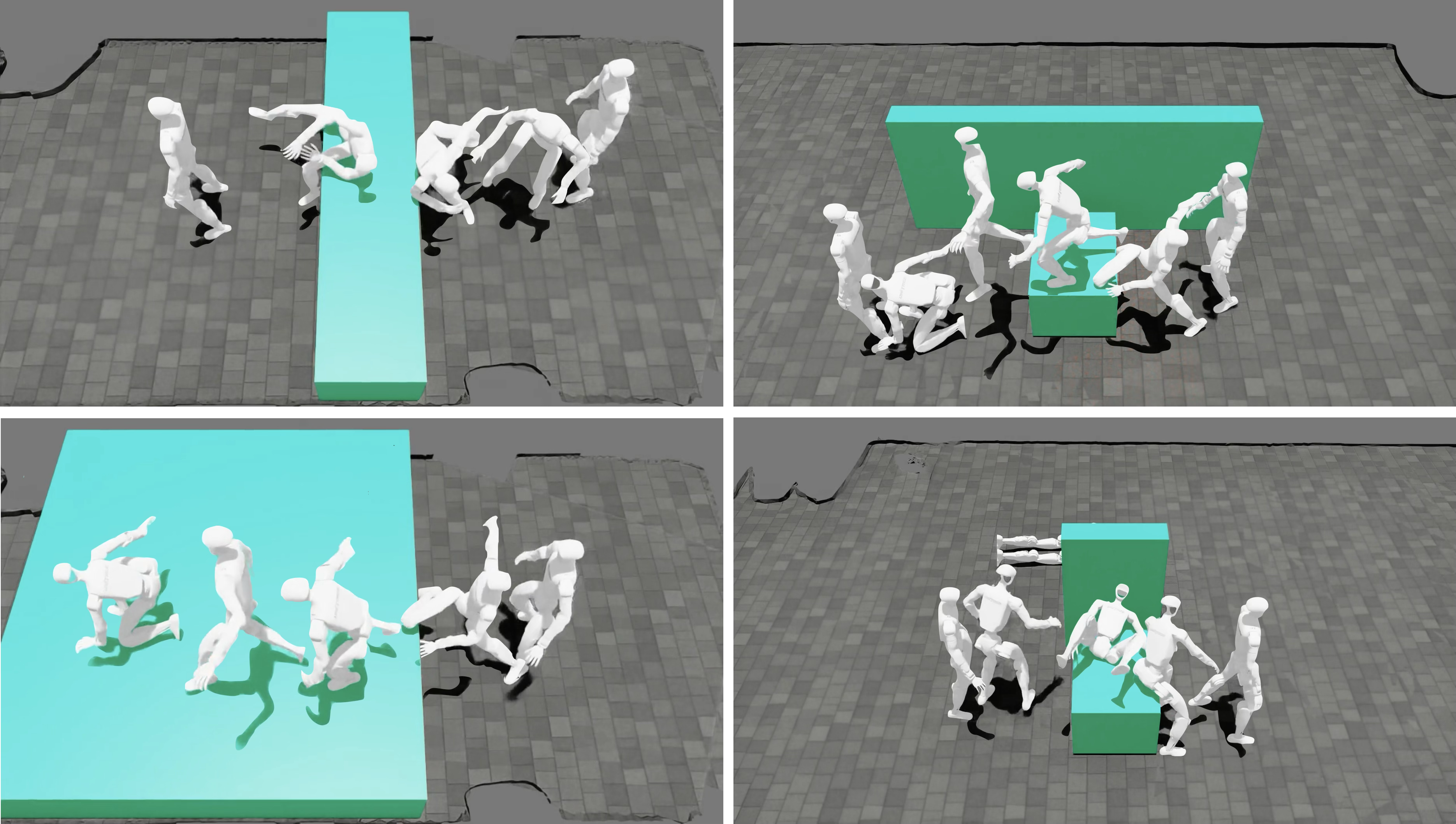}
    \caption{Here we provide examples of adding distractors in the scene to test the robustness of adding unseen objects in the middle of the motion trajectory. We define ``wide distractor'' (top-left), ``plane distractor'' (bottom-left) and ``wall distractor'' (right). They present different visual sensor information while not disrupting the original motion trajectory too much.}
    \vspace{-18pt}
    \label{fig:scene-robustness}
\end{figure}

As shown in Figure~\ref{fig:scene-robustness}, we construct different objects in the scene as distractors and make sure the robot sees the additional objects. Some of the distractors do disrupt the dynamics of the entire motion, e.g, the bottom left example. We tested different variants of distractors and observed no drop in the success rate when running the same motion tracking system.

\begin{table}[h!]
    \centering
    \caption{Motion Tracking Evaluation: MPJPE-Global $\text{MPJPE}_g$ and MPJPE-Base $\text{MPJPE}_b$ across different distractor conditions.}
    \begin{tabular}{lcccc}
        \toprule
        \multirow{2}{*}{\textbf{Distractor Condition}} & \multicolumn{2}{c}{\textbf{Kneel Climb}} & \multicolumn{2}{c}{\textbf{Roll Vault}} \\
        \cmidrule(r){2-3} \cmidrule(l){4-5}
        & $\text{MPJPE}_g$ & $\text{MPJPE}_b$ & $\text{MPJPE}_g$ & $\text{MPJPE}_b$ \\
        \midrule
        No Distractor    & 0.1022 & 0.0390 & 0.1159 & 0.0349 \\
        Wide Distractor  & 0.4848 & 0.0405 & 0.1543 & 0.0373 \\
        Plane Distractor & 0.4261 & 0.0417 & 0.5598 & 0.0809 \\
        Wall Distractor  & 0.1603 & 0.0401 & 0.1157 & 0.0372 \\
        \bottomrule
    \end{tabular}
    \label{tab:mpjpe-vision-robustness}
\end{table}
We focus on 2 motions that are suitable to add distractors while not significantly blocking the dynamics of the motion trajectory. As shown in Table~\ref{tab:mpjpe-vision-robustness}, we test the MPJPE in both the base frame and the world frame compared with the original terrain used in training. In total, there are 4 variants for both roll-vaulting and kneel-climbing. For each variant and the original training scene, we run a batch of 100 robots using the same policy network in the simulator at once. To make a fair comparison, we remove randomization when initializing the robot in the scene. As shown in Table~\ref{tab:mpjpe-vision-robustness}, only the plane distractor significantly increase the MPJPE metric, since the long platform changes the potential motion dynamics compared to the original reference motion. For the wide distractor, it also leads to larger MPJPE. By analyzing the motion, the wide beam disrupts the localization ability of the motion tracking policy. Thus, the MPJPE metric increases.

\subsection{Ablation on the training recipe}
In this section, we run ablation studies on several critical components of the entire training framework. There are several critical components in our training framework:
\begin{enumerate}
    \item Training with depth input.
    \item Randomizing the initialization strategy with stuck detection, which prevents physically impossible rollouts.
    \item Defining local tracking rewards in the relative frame instead of the robot's base frame.
\end{enumerate}
As shown in Table~\ref{tab:mpjpe-ablation}, training motion tracking without exteroception is not robust to initial position perturbation. The starting position has to be exact, or it may face catastrophic failure during scene interaction. When trained only with local frame reward instead of the reward in the relative frame, the policy tends to track more accurate locally, but performs worse than our pipeline globally in interaction intensive cases. Also, we show the non-trivial benefit of applying stuck detection mechanism to prevent large portion of rollout steps becoming useless data samples. Training without stuck detection leads to slightly higher MPJPE error, while still successfully finished the entire motion tracking trajectory.

\section{Conclusion} 
\label{sec:conclusion}
In this work, we present a generalizable paradigm for humanoids to learn agile, physics-based interactions directly from human demonstrations and onboard sensing. We advance beyond standard motion tracking by unifying it with perceptive control into a \textbf{single scalable training framework}. Our analysis highlights the non-trivial benefits of depth vision for robustness against environmental distractors and validates the system through successful deployment in diverse indoor and outdoor real-world scenarios.

Furthermore, by enabling fully end-to-end training with exteroception, this framework provides the critical infrastructure required for scaling up the training system of humanoid whole-body control. This end-to-end approach unlocks the potential to train extensive libraries of motion skills with intensive scene interaction. While challenges regarding data volume and autonomous skill selection remain, they outline a clear path for future research. Ultimately, this work establishes a baseline for developing general-purpose humanoid controllers capable of mastering contact-rich environments in a fully learned, end-to-end manner.

\clearpage

\ifanonsubmission
{} 
\else
{
\section*{Acknowledgments}
This work is accomplished with the help of Xiangting Meng, Siqiao Huang, Yizhuo Gao, and Tairan He.
}
\fi

\bibliographystyle{plainnat}
\bibliography{references}



\end{document}